%% file: root.tex
\documentclass[letterpaper, 10 pt, conference]{ieeeconf}  

\IEEEoverridecommandlockouts                              

\usepackage{xcolor}
\usepackage{amsfonts}

\usepackage{amsmath,amssymb}
\usepackage{microtype} 
\usepackage{balance}
\usepackage{listings}
\usepackage{booktabs}
\usepackage{mathtools}
\usepackage{pxfonts}
\usepackage{graphicx}
\usepackage{multirow}
\usepackage[normalem]{ulem}
\usepackage{fancyvrb}
 \usepackage[ruled,vlined]{algorithm2e}
 \usepackage{algorithmic}

\usepackage{comment}

\newcommand{\tasksymbol}{\Pi}
\newcommand{\vars}{\mathcal{V}}
\newcommand{\var}{v}
\newcommand{\facts}{F}
\newcommand{\ffact}{f}

\newcommand{\states}{S}
\newcommand{\state}{s}
\newcommand{\initstate}{\ensuremath{\state_0}}
\newcommand{\goaldescription}{\ensuremath{\state_{\star}}}

\newcommand{\plan}{\pi}
\newcommand{\actions}{A}

\newcommand{\action}{a}

\newcommand{\eff}{\mathit{eff}}
\newcommand{\pre}{\mathit{pre}}
\newcommand{\perm}{\mathit{u}}

\newtheorem{definition}{Definition}
\newtheorem{proposition}{Proposition}

\usepackage{graphicx}

\title{\LARGE \bf
Understanding a Robot's Guiding Ethical Principles via Automatically Generated Explanations
}

\author{Benjamin Krarup$^{1}$, Felix Lindner$^{2}$, Senka Krivic$^{3}$, and Derek Long$^{1}$
\thanks{$^{1}$ Authors are with the Faculty of Natural and Mathematical Sciences, King's College London, Bush House, WC2B 4BG, London, UK
        {\tt\small firstname.lastname@kcl.ac.uk}}%
\thanks{$^{2}$Felix Lindner is with the Institute of Artificial Intelligence, Ulm University, 89081 Ulm, Germany
        {\tt\small felix.lindner@uni-ulm.de}}%
\thanks{$^{3}$Senka Krivic is with Faculty of Electrical Engineering, University of Sarajevo, 7100 Sarajevo, Bosnia and Herzegovina
        {\tt\small senka.krivic@etf.unsa.ba}}%
}

\begin{document}

\maketitle
\thispagestyle{empty}
\pagestyle{empty}

\begin{abstract}

The continued development of robots has enabled their wider usage in human surroundings.
Robots are more trusted to make increasingly important decisions with potentially critical outcomes.
Therefore, it is essential to consider the ethical principles under which robots operate. 
In this paper we examine how contrastive and non-contrastive explanations can be used in understanding the ethics of robot action plans.
We build upon an existing ethical framework to allow users to make suggestions about plans and receive automatically generated contrastive explanations. Results of a user study indicate that the generated explanations help humans to understand the ethical principles that underlie a robot's plan.

\end{abstract}

\input{introduction}
\input{relatedwork}

\input{planningformalism}
\input{runningexample}
\input{moralprinciples}

\input{ethicalplan}
\input{userstudy}
\input{conclusions}
\bibliographystyle{IEEEtran}
\bibliography{IEEEexample.bib}

\end{document}


\section*{Additional Materials}
\subsection{Soundness Proof}


\begin{proposition}[Soundness]
For all $\Pi$ the HModel $\Pi' = \Pi \times e$ where $e$ is an ethical principle formalised previously it holds that any plan for $\Pi'$ is a plan for $\Pi$.
\end{proposition}

\begin{proof}
For all $\Pi$ where the HModel $\Pi' = \Pi \times \textit{(act-)deontology}$ the proposition holds because the model is only changed by removing actions, therefore $\forall \pi: \rho(\Pi')$ there is a direct mapping of the actions in $\pi$ to actions in $\Pi$, and the goal and initial state are unchanged. Therefore $\pi$ is also a plan for $\Pi$.

For all $\Pi$ where the HModel $\Pi' = \Pi \times \textit{utilitarianism}$ where $\Pi' = \langle \vars, \actions, \initstate, \goaldescription' \rangle$. The actions and initial state are unchanged therefore $\forall \pi: \rho(\Pi')$ there is a direct mapping of the actions in $\pi$ to actions in $\Pi$, and $\goaldescription \subseteq \goaldescription'$ (from line 6 in Algorithm 1). Therefore $\pi$ is also a plan for $\Pi$.

For all $\Pi$ the HModel $\Pi' = \Pi \times \textit{do-no-harm}$ where $\Pi' = \langle \vars', \actions, \initstate', \goaldescription' \rangle$ the actions in $\Pi'$ are the actions in $\Pi$ extended to make the predicate $produced$ true or false. As the predicate $produced$ is not in $\vars$, actions in $\Pi'$ can have $produced$ removed to directly map the action to an action in $\Pi$, $\initstate \subseteq \initstate'$, and $\goaldescription \subseteq \goaldescription'$. Therefore $\forall \pi: \rho(\Pi')$ $\pi$ is also a plan for $\Pi$.
\end{proof}

\begin{proposition}[Soundness]
For all $\Pi$ the HModel $\Pi' = \Pi \times e$ where $e$ is an ethical principle formalised previously it holds that any plan for $\Pi'$ also satisfies $e$.
\end{proposition}

\begin{proof}
For all $\Pi$ the HModel $\Pi' = \Pi \times \textit{(act-)deontology}$ where $\Pi' = \langle \vars, \actions', \initstate, \goaldescription \rangle$, then $\forall a \in A': Good(a) \vee Neutral(a)$. Thus, $\forall \pi: \rho(\Pi')$ which is a sequence of $a \in A'$, $\langle\Pi',\pi\rangle \models \bigwedge_i\lnot Bad(a_i)$ holds.

For all $\Pi = \langle \vars, \actions, \initstate, \goaldescription \rangle$ the HModel $\Pi' = \Pi \times \textit{utilitarianism}$ as the two outputs of Algorithm 1 are $\Pi'$ or $Impermissible$. If the output is $\Pi' = \langle \vars, \actions, \initstate, \goaldescription' \rangle$ then $\goaldescription \subseteq \goaldescription'$ and as per the $max$ function description $GEq(\goaldescription', s_i)$ $\forall s_i \in \mathcal{P}(\vars)$ where $s_i$ is reachable. Therefore because $\goaldescription'$ is well defined so $\forall \pi \in \rho(\Pi')$, $\pi$ must end in $\goaldescription'$, $\langle\Pi',\pi'\rangle \models \bigwedge_iGEq(\goaldescription', s_i)$ holds. If the output is $Impermissible$ then no model $\Pi'$ is returned, therefore the proposition trivially holds.

For all $\Pi$ the HModel $\Pi' = \Pi \times \textit{do-no-harm}$ where $\Pi' = \langle \vars', \actions, \initstate', \goaldescription' \rangle$, $\forall \pi \in \rho(\Pi')$ where $s_n$ is the final state in $\pi$ then $\forall p \in \vars': Bad(p)$ either $\neg p \vee \neg produced_p$ holds in $s_n$. Therefore in $s_n$, $\forall p \in \vars': Bad(p)$ they are either not true, or not caused (because $\neg produced_p \in \initstate', \forall p: Bad(p)$, so if $\exists p \in s_n$: Bad(p) then it was already true and not addded as an effect of an action) so $\langle\Pi',\pi\rangle \models \bigwedge_i (Bad(p_i)\rightarrow \lnot Caused(p_i))$ holds.
\end{proof}

\subsection{[Weak Completeness Proof}

\begin{proposition}[Weak Completeness]
For all $\Pi$ the HModel $\Pi' = \Pi \times e$ where $e$ is an ethical principle formalised previously it holds that if there are no valid plans for $\Pi'$, there are no permissible plans for $\Pi$.
\end{proposition}

\begin{proof}
For each of the ethical principles we prove the contrapositive.


Let $\pi\in\rho(\Pi)$, which is a sequence of $a \in A$, be a permissible plan (according to the (act-)deontology principle). Then $\langle\Pi,\pi\rangle \models \bigwedge_i\lnot Bad(a_i)$ holds. Therefore, this is also a possible plan for $\Pi' = \Pi \times \textit{(act-)deontology}$.


Let $\pi\in\rho(\Pi)$ be a permissible plan (according to the utilitarianism principle). In the final state $s_n$, $\langle \Pi,\pi \rangle \models \bigwedge_iGEq(s_n, s_i)$ holds. The $max$ function returns $s_n$ first in Algorithm 1 and as $\pi$ is a valid plan for $\Pi$ then $\goaldescription \subseteq s_n$ the condition on line 6 of Algorithm 1 will be satisfied. Thus, Algorithm 1 also returns a model $\Pi' \times \textit{utilitarianism}$ that produced a plan that ends in the highest utility state $s_n$.

Let $\pi\in\rho(\Pi)$ be a permissible plan (according to the do-no-harm principle). Then, every fact $p_0, \ldots, p_m$ in the final state $s_n$ is either not bad or not caused. In case that all facts are not bad, then for all bad facts $p$, $\lnot p$ holds in the final state.  Thus, all additional constraints in $\Pi' = \Pi\times \textit{donoharm}$ on the goal are fulfilled. Therefore, $\pi$ is also a plan for $\Pi'$. On the other hand, if there is a bad fact $p$ in the final state, then, as $\pi$ is permissible, $p$ was not caused, that is, no action in $\pi$ has $p$ as an effect. 
Therefore, applying $\pi$ in $\Pi'$ results in $s_n'$, where $\lnot produced_p$ holds for all bad facts.
Again, all additional goal constraints are fulfilled, and thus $\pi$ is a valid plan for $\Pi'$.
\end{proof}

The proofs for Proposition 3 for the (act-)deontology and do-no-harm principle show strong completeness. However, the utilitarianism proof shows weak completeness. Line 7 will not necessarily return the model which gives the same plan as the original model. This could be adapted to return all models that have the highest utility that satisfy the goal, in which case this would become strongly complete.

\begin{proposition}[Weak Completeness]
For all $\Pi$ the HModel $\Pi' = \Pi \times e$ where $e$ is an ethical principle formalised previously it holds that if there are no valid plans for $\Pi'$, there are no permissible plans for $\Pi$.
\end{proposition}

\begin{proof}
For each of the ethical principles we prove the contrapositive.


Let $\pi\in\rho(\Pi)$, which is a sequence of $a \in A$, be a permissible plan (according to the (act-)deontology principle). Then $\langle\Pi,\pi\rangle \models \bigwedge_i\lnot Bad(a_i)$ holds. Therefore, this is also a possible plan for $\Pi' = \Pi \times \textit{(act-)deontology}$.


Let $\pi\in\rho(\Pi)$ be a permissible plan (according to the utilitarianism principle). In the final state $s_n$, $\langle \Pi,\pi \rangle \models \bigwedge_iGEq(s_n, s_i)$ holds. The $max$ function returns $s_n$ first in Algorithm 1 and as $\pi$ is a valid plan for $\Pi$ then $\goaldescription \subseteq s_n$ the condition on line 6 of Algorithm 1 will be satisfied. Thus, Algorithm 1 also returns a model $\Pi' \times \textit{utilitarianism}$ that produced a plan that ends in the highest utility state $s_n$.

Let $\pi\in\rho(\Pi)$ be a permissible plan (according to the do-no-harm principle). Then, every fact $p_0, \ldots, p_m$ in the final state $s_n$ is either not bad or not caused. In case that all facts are not bad, then for all bad facts $p$, $\lnot p$ holds in the final state.  Thus, all additional constraints in $\Pi' = \Pi\times \textit{donoharm}$ on the goal are fulfilled. Therefore, $\pi$ is also a plan for $\Pi'$. On the other hand, if there is a bad fact $p$ in the final state, then, as $\pi$ is permissible, $p$ was not caused, that is, no action in $\pi$ has $p$ as an effect. 
Therefore, applying $\pi$ in $\Pi'$ results in $s_n'$, where $\lnot produced_p$ holds for all bad facts.
Again, all additional goal constraints are fulfilled, and thus $\pi$ is a valid plan for $\Pi'$.
\end{proof}




%% file: introduction.tex
\section{Introduction}

AI Planning systems 
are used in a variety of complex domains to create a sequence of actions known as a plan to achieve a set of goals from an initial state.
Automated planning 
techniques are commonly used in robotics applications for solving complex high-level tasks, for example, tidying up a children's room with a mobile robot~\cite{Krivic-2020-jair} or inspection and maintenance missions for autonomous underwater vehicles (AUVs)~\cite{8594347}.

When autonomous systems are tasked with making decisions with potentially critical outcomes, it is important for them to behave ethically. Take for example the film ``Robot \& Frank''\footnote[1]{https://www.imdb.com/title/tt1990314/}, in which an aging ex-jewel thief, Frank, is bought a domestic robot to care for him. Once Frank realises that the robot does not distinguish between what is considered legal and criminal, or ethical and unethical, Frank uses the robot to help him restart his career as a burglar. There are many ethical quandaries that the human characters are faced with, these may have been better avoided if the robot also understood different ethical standpoints and moral principles. 
Evaluating the ethics of individual plans is necessary for the integration of AI systems in everyday use. However, it is notoriously hard to formally define ethical behavior for autonomous systems \cite{AmodeiEtAl2016}, an instance of the 
value-alignment problem. Thus, it is not enough to pass judgement on automated plans. We approach this problem by allowing users to make suggestions, when they suspect the system could behave more ethically, to produce alternative plans and compare the moralities of each. 


\begin{figure}[t]
    \centering
    \includegraphics[width=1.0\linewidth]{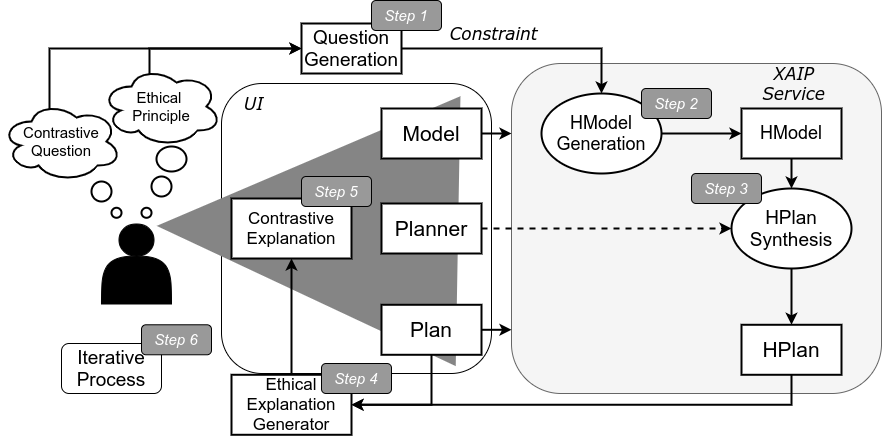}
    \caption{Architecture diagram for the implemented system providing contrastive explanations of the ethics of plans.}
    \label{fig:framework_diagram}
\end{figure}

For example, when Frank is faced with the immanent closure of his love's (Jennifer's) place of work, he decides to stage a heist to win her affection. Frank enlists the help of his robot carer to plan the heist and help him commit it. Under the ethical principle Frank subscribes to, the plan is morally permissible because staging the heist produces the highest utility (impressing Jennifer).
%
However, Frank is unsure about this plan and decides to question the robot.
Explanations to non-contrastive questions such as ``Why are we performing a heist?'', can be useful for understanding the reasoning and intent behind a decision, but do not allow for the exploration of other possibilities.
Frank should also be able to ask contrastive questions of the form ``Why \textit{A} rather than \textit{B}?'', such as ``Why are we performing a heist rather than buying Jennifer flowers?''~\cite{mil18}. These types of questions allow Frank to explore different plans adhering to different principles to better understand 
the ethical consequences of a range of decisions made by the robot.

This paper presents a technical approach to generating non-contrastive and contrastive explanations in the context of ethical AI task planning. We propose a system where a user can ask contrastive questions about the ethics of plans(Figure~\ref{fig:framework_diagram}).
We first outline the planning formalism used throughout the paper and formalise the conditions in which a plan is morally permissible under some principles.
We tailor an approach by Lindnder, Mattmuller, and Nebel~\cite{LindnerEtAl2020} to the definitions of the ethical principles and planning formalism we are concerned with in this paper.
We then demonstrate how to explain the permissibility of plans with automatically generated non-contrastive explanations.
We formalise a set of compilations that \textit{restrict} planning models to ensure that the plans they produce must satisfy certain ethical principles.
We then present an iterative framework for explanation, inspired by Krarup et al.~\cite{kra21}, which allows users to ask contrastive questions about the ethics of plans. Constraints are formed from these questions that produce plans that adhere to the contrast case \textit{B}. Contrastive explanations are generated which show the ethical differences between the original and constrained plan.
We present the results of a user study which shows the effectiveness of the so-generated explanations, mainly, that both non-contrastive and contrastive explanations help people to understand a robot's moral reasoning. However, contrastive explanations are more effective in some circumstances.

This paper is structured as follows: We first present relevant related work in Section~\ref{rel_work}. We introduce the planning formalism and different ethical principles used throughout the paper in Section~\ref{sec:background}. The proposed method for explaining the ethics of a plan is presented in Section~\ref{explain}. Experimental results and discussion are given in Section~\ref{study}. Finally, Section~\ref{concl} concludes the paper. 

%% file: relatedwork.tex
\section{Related Work}
\label{rel_work}

Ensuring autonomous systems work reliably and ethically imposes many research challenges~\cite{den15}. 
There has been recent work on implementing ethical principles for automated moral decision making: Vanderelst and Winfield~\cite{VanderelstWinfield2018} implement a consequentialist ethics by letting a robot simulate different futures and pick the action that leads to the best outcomes. Jackson et al.~\cite{jackson2021integrated} encode moral norms as PDDL3 plan constraints. Lindner and Bentzen~\cite{LindnerBentzen2017hri} employ causal models as representations of available action possibilities. This way, actions can be evaluated with respect to various ethical principles (consequentialists and non-consequentialists).
Research on evaluating the ethics of action plans, rather than single actions, has been conducted by Dennis et al.~\cite{DennisEtAl2016} and Lindner et al.~\cite{LindnerEtAl2020}. Svegliato et al.~\cite{Svegliato_Nashed_Zilberstein_2021} show how ethical principles can be used for ethical sequential decision making by a reinforcement learning agent. 
None of the cited work considers the generation of explanations. However, philosophical work points out that considerations of goodness or choice worthiness in some contexts are better explanations as compared to an agent's motivations or goals \cite{Grimm2016}. Especially when an action plan violates norms, explaining the ethical principles that justify the actions is crucial for understanding.  

There has been a large focus on providing explanations for AI agent behavior recently in the AI community~\cite{cha20} as well as social sciences~\cite{mil17}. 
%
Fox et al.~\cite{fox17} highlighted the usage of contrastive explanations in AI Planning. 
Garfinkel~\cite{gar82} argued that explanations are relative to these contrastive contexts, and that they can be made unambiguous by explicitly stating the contrast case. 
Eifler et al.~\cite{eif20} proposed an approach to answering contrastive questions by citing properties of the plan that would hold if the contrast case were satisfied in the plan.
Krarup et al.~\cite{kra21} provide contrastive explanations by first producing an alternate plan trace including the suggestion made by the user, they then explain the differences between the original and alternate plans. We take inspiration from this to provide contrastive explanations for the ethics of plans.
Autonomous systems need to explain behavior at each level of autonomy and to different individuals~\cite{Setchi20}.
De Graaf and Malle~\cite{8673308} showed that people apply specific explanatory tools for describing autonomous systems compared to other humans, revealing specific
expectations people hold when explaining the behavior of such systems. 
Edmonds~\cite{Edmondseaay4663} claim that components that are best suited to foster trust do not necessarily guarantee best task performance. Also, only a single type of explanation is not sufficient to explain different types of behaviors~\cite{10.1007/978-3-642-16178-0_5}. Therefore, we focused on examining both contrastive and non-contrastive  explanations.

%% file: planningformalism.tex
\section{Background}
\label{sec:background}
We introduce the planning formalism used throughout the paper. We formalise different ethical principles and explain how to use these to ethically validate a plan.

\subsection{Planning Formalism}
We assume that a robot employs a Planning system for sequential decision making. The Planning system's input consists of a planning model formulated in a formal language. The output is a sequence of actions the robot can execute to achieve a given goal.

\subsubsection{Language}
A \emph{planning model} is a tuple $\tasksymbol = \langle
\vars, \actions, \initstate, \goaldescription \rangle$, where $\vars$ is a finite set of Boolean \emph{state variables} $\var$. A \emph{fact} is a state variable or its negation.
The set of all facts is denoted by $\facts$.
A complete conjunction of facts $\state$ is called a
\emph{state}, and $\states$ denotes the set of states of $\tasksymbol$.
The set $\actions$ is a set of \emph{actions}, where an action is a pair
$\action = \langle \pre, \eff \rangle$. The \emph{precondition} $\pre$ and the \emph{effect} $\eff$ are conjunctions of facts.
Every atomic effect may occur at most once in $\eff$.
The state $\initstate \in \states$ is called the \emph{initial state}, and the
partial state $\goaldescription$ specifies the \emph{goal}
condition.

\subsubsection{Semantics}

An action $\action = \langle \pre, \eff \rangle$ is
applicable in state $\state$ iff $\state \models \pre$, i.e., the precondition $\pre$ is satisfied in $\state$. 
Let $\eff =
\bigwedge_{i=1}^{k} v_i$ be an effect, then the \emph{change set} of $\eff$ in $\state$,
symbolically $[\eff]_\state$, is the set of facts $\bigcup_{i=1}^{k} [v_i]_\state$, where $[v_i]_\state = \{v_i \}$ if $\state \models \pre$,
and $\emptyset$, otherwise. 
Applying an action $\action$ to $\state$ yields the
state $\state'$ that has a conjunct $v_i$ for each
$v_i \in [\eff]_\state$, and the conjuncts from $\state$ for all
variables $\var$ that are not mentioned in the change set $[\eff]_\state$. We
write $\state[\action]$ for $\state'$.
We give the following semantics to a sequence of
actions $\plan = \langle
\action_0, \dots, \action_{n-1} \rangle$: 
For $i=0, \dots, n-1$, the next state $\state_{i+1}$ is obtained by 
applying action $\action_i$ to state $\state_i$ (assuming that it is
applicable).
If $\action_i$ is inapplicable in $\state_i$ for some $i=0,\dots,n-1$, then $\plan$ is inapplicable in $\initstate$.
A state $\state$ is a goal state if $\state \models \goaldescription$.
We call $\plan$ a \emph{plan} for $\tasksymbol$ if it is
applicable in $\initstate$ and if its final state $\state_n$ is a goal
state, i.e., $\state_n \models \state_\star$. We use the function $\rho(\Pi)$ to denote the set of all plans for a model $\Pi$ for which the last condition holds.

\subsubsection{Moral valuations of actions and consequences}

Each planning model
$\tasksymbol$ comes with a utility function $\perm\colon A \cup
\facts \to \mathbb{R}$ that maps
 actions and facts to utility values.
An action
$\action$ or fact $\ffact$ is
\emph{morally bad} if $\perm(\action) < 0$ or $\perm(\ffact) < 0$,
respectively. Similarly, an action or fact is
\emph{morally neutral} or
\emph{morally good} if its utility value is zero or greater than zero, respectively.
We explicitly do
\emph{not} require that moral values of actions and facts must be consistent
in any particular sense. For instance, we do not require that
an action must be classified as morally bad if one (or all) of
its effects are morally bad. The rationale behind this choice is that,
under some ethical principles, actions are good or bad \emph{per se}, without regard
to their actual effects. 
When using a consequentialist view, we  judge the moral value of a
plan based on the value of its final state, which is defined to be
the sum over the utility values  of all facts in the final state:
$u(s) = \sum_{\{v\,|\, s\models v\}} \perm(v)$.

%% file: moralprinciples.tex
\subsection{Moral Principles}\label{sec:moral_principles}

\subsubsection{Moral Action Query Language}


Let $V^* = \{p, \lnot p \,|\, p \in labels(A)\cup V\}$ be the set of literals denoting actions and facts and their negations. Language $L$ is the smallest set, such that:
\begin{itemize}
\item For all facts $p\in V^*$ formulae $Caused(p)$ are in $L$.
\item For all actions and facts $p\in V^*$ formulae $Good(p)$, $Bad(p)$, and $Neutral(p)$ are in $L$.
\item For all conjunctions of facts, $p_1\land\ldots\land p_n$, $q_1\land\ldots\land q_m$ formulae $GEq(p_1\land\ldots\land p_n, q_1\land\ldots\land q_m)$ are in $L$.
\item If $\phi,\psi \in L$, then $\lnot \phi, \phi\land \psi, \phi\lor \psi, \phi \rightarrow \psi \in L$.
\end{itemize}

The semantics of $L$ is defined over pairs of planning models and plans $\langle \Pi, \pi \rangle$ as follows:
\begin{itemize}
\item $\langle \Pi, \pi \rangle \models Good(p)$ iff $u(p) > 0$.
\item $\langle \Pi, \pi \rangle \models Bad(p)$ iff $0 < u(p)$.
\item $\langle \Pi, \pi \rangle \models Neutral(p)$ iff $0 = u(p)$.
\item $\langle \Pi, \pi \rangle \models GEq(s, s')$ iff $u(s) \geq u(s')$.
\item $\langle \Pi, \pi \rangle \models \lnot\phi$ iff $\langle \Pi, \pi \rangle \not\models \phi$.
\item $\langle \Pi, \pi \rangle \models \phi\land\psi$ iff $\langle M, w_\alpha \rangle\models \phi$ and $\langle \Pi, \pi \rangle\models\psi$.
\item $\langle \Pi, \pi \rangle \models \phi\lor\psi$ iff $\langle \Pi, \pi \rangle\models \phi$ or $\langle \Pi, \pi \rangle\models\psi$.
\item $\langle \Pi, \pi \rangle \models \phi\rightarrow\psi$ iff $\langle \Pi, \pi \rangle\not\models\phi$ or $\langle \Pi, \pi \rangle\models\psi$.
\item $\langle \Pi, \pi \rangle \models Caused(p)$ iff $p$ holds in the final state of $\pi$, and some action in $\pi$ had $p$ as an effect. 
\end{itemize}

\subsubsection{Ethical Principles Formalized}~\label{sec:ethical_principles_formalised}

Following prior work by  \cite{LindnerEtAl2020}, we formalize conditions under which an action plan counts as morally permissible according to some ethical principle. We consider three ethical principles: \mbox{(act-)deontology}, utilitarianism, and a do-no-harm principle.

For a plan to be permissible w.r.t. \mbox{(act-)deontology}, the plan must not contain any inherently bad actions. 

\begin{definition}
\label{def:deon}
A plan $\pi = a_0\ldots a_{n-1}$ for a planning model $\Pi$ is morally permissible according to \emph{\mbox{(act-)deontology}} if and only if  $\langle\Pi,\pi\rangle \models \bigwedge_i\lnot Bad(a_i)$ holds.
\end{definition}

For a plan to be permissible w.r.t. utilitarianism, the plan must lead to one of the best reachable world states.
\begin{definition}
\label{def:util}
A plan $\pi = a_0\ldots a_{n-1}$ for a planning model $\Pi$, which ends in final state $s_n$ is morally permissible according to \emph{utilitarianism} if and only if $\langle\Pi,\pi\rangle \models \bigwedge_iGEq(s_n, s_i)$, where $s_i$ is a reachable state.
\end{definition}

A plan is said to be permissible w.r.t. the do-no-harm principle, if the plan is not the
cause of some bad outcome.

\begin{definition}
\label{def:dnh}
Let $\Pi$ be a planning task and $\pi$ a plan resulting in final state $s_n$, where facts $p_0,\ldots,p_m$ hold. The plan $\pi$ is morally permissible according to the \emph{do-no-harm principle} if and only if $\langle\Pi,\pi\rangle \models \bigwedge_i (Bad(p_i)\rightarrow \lnot Caused(p_i))$.
\end{definition}

\section{Explaining Robot Action Plans} \label{explain}

In this section, we present 
a method for explaining the ethics of a plan per se (\emph{non-contrastive explanation}), which is then used by a method for generating explanations  with reference to an alternative plan (\emph{contrastive explanation}).

\subsection{Non-Contrastive Moral Explanations} \label{subsec:moralexpl}

We extend the work reported by Lindner and Moellney~\cite{LindnerMoellney2019} to explaining the moral permissibility of plans of actions. This is done by computing the prime-implicants and/or prime-implicates of the formulae that represent the ethical principles. Here, we briefly reproduce the approach by example: Consider a plan $\pi = a_1a_2a_3$ and assume that $a_1$ is a good action whereas $a_2$ and $a_3$ are bad actions. To check if $\pi$ is morally permissible according to the deontology principle, one has to evaluate if $\langle\Pi,\pi\rangle \models \lnot Bad(a_1)\land \lnot Bad(a_2)\land \lnot Bad(a_3)$ holds.
This check fails because $a_2, a_3$ are bad. To automatically generate the reason why the plan is impermissible, the minimal conflict sets are computed, i.e., the minimal sets of literals which entail the falsity of the formula. In the example these are $\{Bad(a_1)\}$, $\{Bad(a_2)\}$, and $\{Bad(a_3)\}$. Each of these sets is a \emph{possible sufficient reason} for the impermissibility of the plan. We filter for those sets which hold in the actual situation, i.e., $\{Bad(a_2)\}$, and $\{Bad(a_3)\}$, and call them \emph{sufficient reasons}. The system can thus pick one reason, e.g., $Bad(a_2)$, and state that the plan is morally impermissible because action $a_2$ is morally bad. However, the badness of $a_2$ is not necessary for the impermissibility of $\pi$, because $a_3$ is bad as well. The \emph{necessary reasons} for the impermissibility of $\pi$ is obtained by computing the minimal hitting set of the sufficient reasons. In the example, this is $\{Bad(a_2), Bad(a_3)\}$. Hence, the system can state that the plan is bad, because $a_2$ and $a_3$ are bad, and thereby suggest to make the plan permissible by avoiding both these facts.

%% file: ethicalplan.tex
\subsection{Contrastive Moral Explanations}
\label{subsec:contrastive_expl}
The method described in the previous section enables us to compute explanations for why a particular plan is permissible or not, but it does not give us a means to come up with an alternative plan for comparison. We formalise an approach to ethical plan generation which will be used in an iterative process for contrastive explanations.

\subsubsection{Ethical Plan Generation} \label{sec:ethical_plan_generation}
We utilise the definition of \textit{Model Restrictions} through the use of the \textit{Constraint Operator} proposed by Krarup et al.~\cite{kra21} to restrict the behaviour of plans to adhere to ethical principles, defined as follows.

A \textit{constraint property} is a predicate, $\phi$, over plans. A \textit{constraint operator}, $\times$ is defined so that, for a planning model $\Pi$ and any constraint property $\phi$, $\Pi \times \phi$ is a model (an \textit{HModel}), $\Pi'$, called a \textit{model restriction} of $\Pi$, satisfying the condition that any plan for $\Pi'$ is a plan for $\Pi$ that also satisfies $\phi$. A plan for an HModel is refered to as an \textit{HPlan}.
We formalise the implementation of the constraint operator for three constraint properties, the (act-)deontology, utilitarianism, and do-no-harm principles defined previously. The specifics of these compilations are as follows.

\paragraph{(Act-)Deontology Principle}
For a planning model $\Pi$ and the ethical principle \textit{e = (act-)deontology}, $\Pi \times e = \Pi'$ where $\Pi' = \langle \vars, \actions', \initstate, \goaldescription \rangle$ and $\actions' = \{\forall a \in \actions | Good(a) \vee Neutral(a)\}$.

\paragraph{Utilitarianism Principle}
For a planning model $\Pi$ and the ethical principle \textit{e = utilitarianism}, $\Pi \times e = \Pi'$ where $\Pi'$ is found through the process described in Algorithm~\ref{alg:algorithm1}. 

\begin{algorithm}[t]
\textbf{Input}: $\Pi = \langle \vars, \actions, \initstate, \goaldescription \rangle$\\
\textbf{Output}: $(\Pi' = \Pi \times$ \textit{utilitarianism)} or \textit{Impermissible}\\
\begin{algorithmic}[1]
 \STATE $powerV = \mathcal{P}(\vars)$;
 \STATE $\goaldescription' = max(powerV, \goaldescription)$;
 \STATE $\Pi' = \langle \vars, \actions, \initstate, \goaldescription' \rangle$;
 \WHILE{$powerV \neq \varnothing$}
  \STATE $\pi = Planner(\Pi')$;
  \IF{$\goaldescription \subseteq \goaldescription' \wedge \pi \neq \varnothing$}
    \RETURN $\Pi'$;
  \ELSIF{$\pi \neq \varnothing$}
    \RETURN $Impermissible$;
  \ENDIF
  \STATE $powerV = powerV \setminus \goaldescription'$;
  \STATE $\Pi' = \langle \vars, \actions, \initstate, \goaldescription' \rangle$;
  \STATE $\goaldescription' = max(powerV, \goaldescription)$;
 \ENDWHILE
 \RETURN $Impermissible$;
\end{algorithmic}
 \caption{Utilitarianism Search.}
 \label{alg:algorithm1}
\end{algorithm}
We assume the $max$ function takes a set of states and returns the highest utility set of state variables i.e. for the set of states $S$ it returns $s_n \in S$ where $GEq(s_n, s_i)$ $\forall s_i \in S \setminus s_n$. If there is more than one set with the same utility the sets containing the original goal set are returned first. The $Planner$ function takes a planning model $\Pi$ and produces either a plan $\pi$ or the empty set if the model is unsolvable.

\paragraph{Do-No-Harm Principle}
For a planning model $\Pi$ and the ethical principle \textit{e = {\mbox{do-no-harm}}}, $\Pi \times e = \Pi'$ where $\Pi' = \langle \vars', \actions, \initstate', \goaldescription' \rangle$ where 
\begin{itemize}
    \item $\vars' = \vars \cup \{produced_p | \forall p \in \vars: Bad(p)\}$
    \item $\initstate' = \initstate \cup \{\neg produced_p | \forall p \in \vars : Bad(p)\}$
    \item $\goaldescription' = \goaldescription \cup \{\lnot p \vee \neg produced_p | \forall p \in \vars: Bad(p)\}$
\end{itemize}
And $\forall a = \langle \pre, \eff \rangle \in \actions | \exists p \in \eff : Bad(p)$, the effects of the actions are extended to make $produced_p$ true, and $\forall a = \langle \pre, \eff \rangle \in \actions | \exists \neg p \in \eff : Bad(p)$, the effects of the actions are extended to make $produced_p$ false.


\begin{proposition}[Soundness]
For all $\Pi$ the HModel $\Pi' = \Pi \times e$ where $e$ is an ethical principle formalised previously it holds that any plan for $\Pi'$ is a plan for $\Pi$.
\end{proposition}

\begin{proposition}[Soundness]
For all $\Pi$ the HModel $\Pi' = \Pi \times e$ where $e$ is an ethical principle formalised previously it holds that any plan for $\Pi'$ also satisfies $e$.
\end{proposition}


From Propositions 1 and 2 the proposition ``For all $\Pi$ the HModel $\Pi' = \Pi \times e$, if there are no permissible plans for $\Pi$ then there are no valid plans for $\Pi'$'' trivially follows.

\begin{proposition}[Weak Completeness]
For all $\Pi$ the HModel $\Pi' = \Pi \times e$ where $e$ is an ethical principle formalised previously it holds that if there are no valid plans for $\Pi'$, there are no permissible plans for $\Pi$.
\end{proposition}

\subsubsection{Contrastive Explanations as an Iterative Process}

We treat explanation as a form of dialogue, more specifically, as an iterative process in which the user asks contrastive questions (``Why \textit{A} rather than \textit{B}'')~\cite{mil19} about the plans produced by robot agents. To produce contrastive explanations it must be possible to reason about the hypothetical alternative (\textit{contrast case}) \textit{B}, which we approach by constructing an alternative plan where \textit{B} is satisfied rather than \textit{A}.

We use the trolley problem to exemplify our approach. Imagine a conductor is directing trains to different tracks through the use of a lever. At a railway junction there are five people stuck on the currently assigned track, and one person on the other; with a train about to reach the junction. The conductor can choose to pull a lever to send the train down the other track killing the one person, if they do nothing, then the five people will die.

We can model the trolley problem as a planning model 
$\tasksymbol = \langle \vars, \actions, \initstate, \goaldescription\rangle$, with $\vars = \{5willdie, 1willdie, done\}$, $\actions = \{pull,\mathit{refrain}\}$, $\initstate = \{5willdie,\lnot 1willdie,\lnot done\}$, $\goaldescription = \{done\}$, and $pull = \langle \top, \lnot 5willdie\land 1willdie\land done\rangle$, $\mathit{refrain} = \langle\top, done\rangle$. The moral valuation function assigns $u(5willdie) = -5$, $u(\lnot 5willdie) = 5$, $u(1willdie) = -1$, $u(\lnot 1willdie) = 1$, $u(done) = u(\lnot done) = u(pull) = u(\mathit{refrain}) = 0$. There are two (shortest) plans that reach the goal: $\plan = \mathit{refrain}$ and $\plan' = pull$. Plan $\plan$ is impermissible from the utilitarian point of view because $\plan'$ leads to a better state. From the deontological point of view, both $\plan$ and $\plan'$ are permissible because they do not contain inherently bad actions. The do-no-harm principle renders $\plan$ permissible (as the only bad effect $5willdie$) is not produced by $\plan$; and it renders $\plan'$ impermissible because the bad effect $1willdie$ is produced by $pull$.



In our example, the conductor is using an AI Planning System (AIPS) used within the framework described in Figure~\ref{fig:framework_diagram} to help them manage the complex rail network. Consider the case where the AIPS is behaving under the do-no-harm principle, so the plan that is presented to the conductor is ``refrain from pulling the lever''. The conductor does not understand why the AIPS would allow five people to be killed rather than one. 
\textit{(Step 1)} The conductor asks the contrastive question ``Why did you refrain from pulling the lever, rather than pulling it?'', and provides the ethical principle for which they want to understand, in this case the do-no-harm principle. 
\textit{(Step 2)} The \textit{Question Generation} module extracts a constraint from the contrast case given by the user's question and the ethical principle, specifically that the lever is pulled and the do-no-harm principle is satisfied. We then use the same definition of the constraint operator to restrict the model with this constraint to produce the HModel $\Pi'$. The questions we provide contrastive explanations for, how the constraints are derived, and how they are applied is described in detail by Krarup et al.~(\cite{kra19,kra21}).
These questions are restricted to forms where $A$ and $B$ are properties on the actions in the plan. More specifically questions about inclusion, exclusion, or orderings of actions in the plan, for example, ``why was action $a_1$ used before action $a_2$?''. 
\textit{(Step 3)} The HModel is solved with the planner used within the AIPS to produce an HPlan, which satisfies the constraint derived from the conductors question and ethical principle.
If the ethical principle cannot be satisfied (the HModel is unsolvable), then there will be no HPlan (from propositions 1 and 2). The ethical principle constraint is removed from the HModel and it is re-solved so that an HPlan can be found and used in the next step. The explanation will then show why the user's contrast case leads to a plan that is impermissible.
\textit{(Step 4)} The original plan and the HPlan are passed to the \textit{Ethical Explanation Generator} which returns the necessary reason for the plan being (im-)permissible, this is the non-contrastive moral explanation. In this case it will return \textit{Permissible}: $\lnot Caused(\mathit{5willdie})$ for the original plan and \textit{Impermissible}: $Caused(\mathit{1willdie})$ for the HPlan. 
\textit{(Step 5)} A \textit{Contrastive Explanation} is generated which aims to show the difference between these reasons. This is formed from the set difference of the two reasons;
and the actions in the plan, the permissibility, principle, and the cause, contrasted to the same information for the HPlan.
For example, the contrastive explanation produced to answer the conductor's question is ``The man could refrain from action. This would be permissible under the do-no-harm principle because this way the death of the five persons is not caused by his action. Alternatively, the man could pull the lever. Doing so is impermissible under the do-no-harm principle because this way the death of the one person is caused by his action.''
In this example the HModel $\Pi'$ is unsolvable because any plan where the lever is pulled is impermissible under the do-no-harm principle. Therefore, as said in Step 4, the permissibility constraint is dropped so that the HModel can be solved to produce an HPlan where the lever is pulled. This HPlan is used in Step 4 to generate the reasons for why it is impermissible, ``...because this way the death of the one person is caused by his action''. 
\textit{(Step 6)} This whole process can then be iterated, where a user can ask new questions and explore different ethical principles and outcomes until they are satisfied that they either understand the ethics of the plan well, or they have found a plan that meets their moral standards.
For example, the conductor still might not be satisfied with the choice to refrain from pulling the lever. The conductor asks the same question as above but instead chooses the utilitarianism principle. Again, the constraint will ensure that the lever is pulled in the plan, but now, that the utilitarianism principle is satisfied. The necessary reasons will again be \textit{Permissible}: $\lnot Caused(\mathit{5willdie})$ for the original plan and \textit{Permissible}: $GEq(\mathit{1willdie} \land \lnot \mathit{5willdie}\land done, \lnot \mathit{1willdie} \land \mathit{5willdie}\land done)$ for the new HPlan. The contrastive explanation will then be ``The man could refrain from action. This would be permissible under the do-no-harm principle because this way the death of the five persons is not caused by his action. Alternatively, the man could pull the lever. Doing so is permissible under the utilitarianism principle because because five saved lives is better than one saved life.'' The conductor agrees with the outcome of the utilitarianism principle and decides to execute the plan of pulling the lever. 
Explanations produced in this type of iterative framework have been shown to help users understand plans better~\cite{kra21}. However, it is not known if the contrastive and non-contrastive explanations discussed in this paper help users to understand ethical principles better, and which are more effective. We performed a user study to test these hypotheses.

%% file: userstudy.tex
\section{Empirical Evaluation of the Generated Explanations}
\label{study}

We have conducted a between-subject online-questionnaire study with six conditions, \{Deontology, Utilitarianism, DoNoHarm\}~$\times$~\{Contrastive, NonContrastive\} to test whether our explanations support humans to build an understanding of the robot's moral reasoning. Moreover, we were interested in investigating whether the explicit mentioning of the alternative plan in the verbalization of the contrastive explanations has a positive effect. The hypotheses read:
(\textbf{H1}) The explanations have an effect on people's understanding of the robot's moral reasoning; (\textbf{H2}) Contrastive explanations are more effective in enabling people to build a mental model of the robot's moral reasoning as compared to non-contrastive explanations.


\subsection{Methods}

\subsubsection{Materials}

The first stage of the online questionnaire asked participants for age and gender. In the second stage, the classical trolley problem was presented, both as an image and as text:
\textit{``There is a runaway trolley barrelling down the railway tracks. Ahead, on the tracks, there are five people tied up and unable to move. The trolley is headed straight for them. A man is standing some distance off in the train yard, next to a lever. If he pulls this lever, the trolley will switch to a different set of tracks. However, he notices that there is one person on the side track. He has two options: Do nothing and allow the trolley to kill the five people on the main track. Pull the lever, diverting the trolley onto the side track where it will kill one person. 
What is the right thing to do?''}

Next, the moral reasoning robot was introduced to the participant:
\textit{``The robot makes moral judgments based on a moral principle implemented into the robot operating system. People can take recommendations based on these judgments. The robot's moral judgment reads: ...''}

Depending on the conditions, the sentence continued:

\textit{Deontology, Non-Contrastive}: {``The man can pull the lever. Doing so is permissible because pulling a lever is not inherently bad.''}

\textit{Deontology, Contrastive}: {``The man can pull the lever. Doing so is permissible because pulling a lever is not inherently bad. Alternatively, the man can refrain from pulling the lever. This is also permissible, because doing nothing is not inherently bad.''}

\textit{Utilitarianism, Non-Contrastive}: {``The man can pull the lever. Doing so is permissible because five saved lives is better than one saved life.''}

\textit{Utilitarianism, Contrastive}: {``The man can pull the lever. Doing so is permissible because five saved lives is better than one saved life. Alternatively, the man can refrain from pulling the lever. This would be impermissible, because one saved life is worse than five saved lives.''}

\textit{Do-No-Harm, Non-Contrastive}: {``The man can pull the lever. Doing so is impermissible because this way the death of the one person is caused by the man's action.''}

\textit{Do-No-Harm, Contrastive}: 
    {``The man could pull the lever. Doing so is impermissible because this way the death of the one person is caused by his action. Alternatively, the man could refrain from action. This would be permissible because this way the death of the five persons is not caused by his action.''}

After the presentation of the robot's judgment, participants were asked to check a checkbox to confirm that they have read the judgment: \textit{Have you read and understood the robot's judgment?}

The third stage was designed as a prediction task to test the participant's mental model of the robot's moral reasoning.
The bridge dilemma, a variation of the trolley problem, was presented:
\textit{``A trolley is hurtling down a track towards five people. A man is standing on a bridge under which it will pass, he can stop it by putting something very heavy in front of it. There is a very fat man next to him -- his only way to stop the trolley is to push that man over the bridge and onto the track, killing him to save five.
What is the right thing to do?''}

The text was accompanied by an image showing the bridge dilemma. Next, participants were asked for their prediction:
\textit{
    ``Applying what you already know about the robot‘s reasoning from the previous moral judgement of the trolley problem, irrespective of your own moral judgment, rate how likely you consider the robot to come to each of the following moral judgments for this new dilemma.''}
The available options were:
\textit{``It is ethical to push the man off the bridge.''},
\textit{``It is ethical to refrain from pushing the man off the bridge.''},
\textit{``Both options are ethical.''}, and 
\textit{``None of the options are ethical.''}.
For each option, the participant gave a likeliness rating on a 5-point Likert scale: Very Unlikely---Rather Unlikely---Undecided---Rather Likely---Very Likely.

\subsubsection{Procedure}
Participants were recruited via Prolific to participate in the online survey ``Moral Reasoning in Robots''. Completion of the questionnaire took about 10 minutes.

\subsubsection{Participants}

We recruited $N=300$ participants. To ensure quality of data, we removed all data points where the participants needed less than 3 minutes or more than 30 minutes to complete the questionnaire. The 3-minutes threshold is justified by the fact that even just reading through the questionnaire takes three minutes. Thus, it is very likely that the participants did not take the questionnaire seriously. The 30-minutes threshold is justified by the fact that even if one is a very slow reader and thinking about every detail, the questionnaire should not take more than that.
The remaining $N=245$ data points where used for analysis. The mean age of participants was 27 (SD = 10), 82 were female, 159 were male, and 4 did not specify gender. Each participant filled out one of six questionnaires. From the 245 participants that were included into the analysis, 41 were assigned to the non-contrastive deontological explanation, 37 to the contrastive deontological explanation, 40 to the non-contrastive do-no-harm explanation, 44 to the contrastive do-no-harm explanation, 39 to the non-contrastive utilitarian explanation, and 41 to the contrastive utilitarian explanation.
All participants got $\pounds1.50$ for an estimated effort of 10 minutes to complete the questionnaire.

\begin{figure*}[t]
    \centering
    \includegraphics[width=\textwidth]{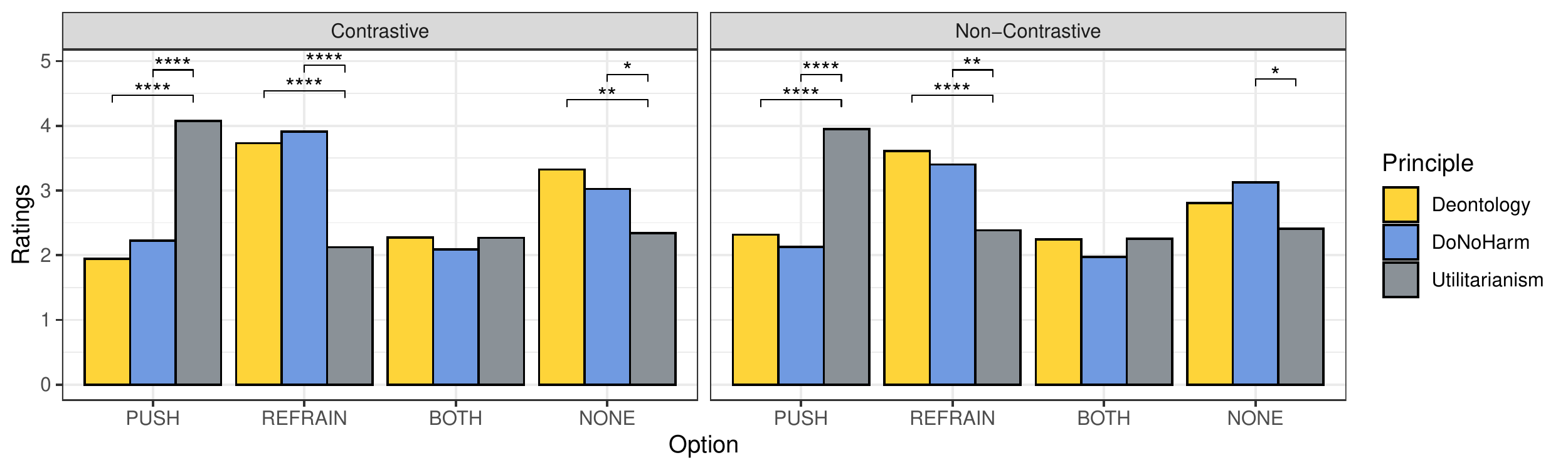}
    \caption{Mean permissibility ratings per course of action for all six conditions.  $ {}^* p \leq .05, {}^{**} p\leq .01, {}^{***} p\leq .001$}
    \label{fig:barplot}
\end{figure*}
\subsection{Results}

A three-way ANOVA was run to examine the influence of the variables Option (= the option for acting), Principle (= the principle the explanation is based on) and Contrast (= the form of the explanation, viz., whether it is contrastive or non-contrastive) on the permissibility ratings. There was a significant interaction between Option and Principle ($F(6, 944) = 33.452$, $p < 0.0001$). That is, the ratings that each option receives is significantly depending on the explanation's content.
In further support of hypothesis H1, post-hoc pairwise Bonferroni-adjusted comparisons show significant influences of the various explanations on the permissibility ratings (see Fig.\ \ref{fig:barplot}):

Participants who had read the deontological explanation rated the option that pushing is permissible lower as compared to participants who had read the utilitarian explanation. The difference is significant for contrastive ($F(1, 944) = -7.44$, $p < .0001$) and for non-contrastive explanations ($F(1, 944) = -5.79$, $p < .0001$). 
The deontological explanation group also rated the option that refraining is permissible higher as compared to the utilitarian explanation group. The difference is significant for contrastive ($F(1, 944) = 5.62$, $p < .0001$) and for non-contrastive explanations ($F(1, 944) = 4.34$, $p < .0001$). Moreover, deontological explanations lead to higher ratings for the option that neither pushing nor refraining is permissible as compared to utilitarian explanations. The difference is significant for contrastive ($F(1, 944) = 3.44$, $p < .001$) but not for non-contrastive explanations ($F(1, 944) = 1.4$, $p < .486$). 

Participants who had read the do-no-harm explanation rated the option that pushing is permissible lower as compared to  participants who had read the utilitarian explanation. The difference is significant for contrastive ($F(1, 944) = -6.74$, $p < .0001$) and for non-contrastive explanations ($F(1, 944) = -6.43$, $p < .0001$). 
The do-no-harm explanation group also rated the option that refraining is permissible higher as compared to the utilitarian explanation group. The difference is significant for contrastive ($F(1, 944) = 6.53$, $p < .0001$) and for non-contrastive explanations ($F(1, 944) = 3.58$, $p < .001$). Moreover, do-no-harm explanations lead to higher ratings of the option that neither pushing nor refraining is permissible higher as compared to utilitarian explanations. The difference is significant for contrastive ($F(1, 944) = 2.49$, $p = 0.039$) and for non-contrastive explanations ($F(1, 944) = 2.52$, $p = 0.036$). 

Regarding hypothesis H2, we observe that the influence of the contrastive explanation is generally more pronounced than of the non-contrastive explanation. However, the difference between contrastive and non-contrastive explanations does not reach statistical significance. Most notably, we find that the contrastive deontological explanation group is more convinced that nothing is permissible in the bridge dilemma than the non-contrastive deontological explanation group ($F(1,944) = 1.82, p = .07, n.s.$), and that the contrastive do-no-harm explanation group considers refraining more permissible than the non-contrastive do-no-harm explanation group ($F(1, 944) = 1.85$, $p=.065, n.s.$).

\subsection{Discussion}

\paragraph{Mental Models} The results support hypothesis H1 as they indicate that the explanations enabled people to build a mental model of the robot's moral reasoning.
In particular, the results rule out the possibility that the participants relied on their own moral view as the ratings for permissibility of the available options differed significantly. Note that also the sentence ``Doing so is (im-)permissible...'' cannot explain these differences because in our version of the trolley dilemma, pulling the lever is permissible under the deontic and the utilitarian principle but in the bridge dilemma, deontology forbids and utilitarianism permits pushing the man off the bridge. Participants clearly noticed what the robot considers permissible in each condition. The only explanation for this is that participants took notice of the explanations and transferred their understanding of the robot's explanation of its moral reasoning in the trolley problem to the bridge dilemma. 

A limitation of the conducted study is that this result was obtained by testing a small number of principles applied to one moral dilemma. More work is needed in understanding to what extent the result can be generalized to more complicated moral principles and dilemmas. More complicated moral principles consist of multiple rules (e.g., the Doctrine of Double Effect \cite{Bentzen2016}) or require universalization (e.g., the first formulation of Kant's Categorical Imperative \cite{Powers2006}). To be understandable then, it may be required that explanations also contain an explicit explanation of the moral principle used.  
More complicated dilemmas would consist of sequences of actions over longer temporal horizons and with more complex causal interdependencies.
\paragraph{Non-contrastive vs. contrastive explanations} Regarding the comparison between non-contrastive and contrastive explanations, the results are less supportive of H2. 
One explanation for this may be that humans have the disposition to draw contrastive inferences all the time (cf., \cite{mil19}). Thus, the participants in the non-contrastive conditions may have been able to infer the contrastive part of the explanation themselves. 
An alternative explanation for the small effect of contrastivity may be that the information contained in the non-contrastive explanation was already sufficient for understanding the ethical principle. That is, the non-contrastive deontology group inferred that an action is impermissible if inherently bad, and pushing someone is inherently bad. The non-contrastive utilitarian group also had no problem inferring that permissibility is grounded in counting saved lives. Only the non-contrastive do-no-harm group was less able (as compared to the contrastive do-no-harm group) to infer from their explanation that refraining from action is permissible. This information is less obvious and only contained in the contrastive explanation. 
We leave the design of another user study to test this new hypothesis for future work.
Generally, more research is needed to understand which inferences people automatically draw from explanations and under which circumstances explicit contrastivity provides additional information.

%% file: conclusions.tex
\section{Conclusions}
\label{concl}
We have presented a system that enables autonomous systems to explain action plans in terms of ethical principles. Our explanation method compiles a user's why-question into a new AI planning problem. 
The technique was integrated into a system that allows humans to actively explore the explanatory space of ethical and unethical alternative plans. Our user study indicates that the  explanations generated by the system can improve humans' understanding of a robot's ethical principle. As well as further studies, future work will also investigate the compilation of more intricate ethical principles into commonly used planning formalisms.